# LexRank: Graph-based Lexical Centrality as Salience in Text Summarization

**Güneş Erkan**                                                          GERKAN@UMICH.EDU
*Department of EECS*
*University of Michigan, Ann Arbor, MI 48109 USA*

**Dragomir R. Radev**                                                    RADEV@UMICH.EDU
*School of Information & Department of EECS*
*University of Michigan, Ann Arbor, MI 48109 USA*

## Abstract

We introduce a stochastic graph-based method for computing relative importance of textual units for Natural Language Processing. We test the technique on the problem of Text Summarization (TS). Extractive TS relies on the concept of sentence salience to identify the most important sentences in a document or set of documents. Salience is typically defined in terms of the presence of particular important words or in terms of similarity to a centroid pseudo-sentence. We consider a new approach, LexRank, for computing sentence importance based on the concept of eigenvector centrality in a graph representation of sentences. In this model, a connectivity matrix based on intra-sentence cosine similarity is used as the adjacency matrix of the graph representation of sentences. Our system, based on LexRank ranked in first place in more than one task in the recent DUC 2004 evaluation. In this paper we present a detailed analysis of our approach and apply it to a larger data set including data from earlier DUC evaluations. We discuss several methods to compute centrality using the similarity graph. The results show that degree-based methods (including LexRank) outperform both centroid-based methods and other systems participating in DUC in most of the cases. Furthermore, the LexRank with threshold method outperforms the other degree-based techniques including continuous LexRank. We also show that our approach is quite insensitive to the noise in the data that may result from an imperfect topical clustering of documents.

## 1. Introduction

In recent years, natural language processing (NLP) has moved to a very firm mathematical foundation. Many problems in NLP, e.g., parsing (Collins, 1997), word sense disambiguation (Yarowsky, 1995), and automatic paraphrasing (Barzilay & Lee, 2003) have benefited significantly by the introduction of robust statistical techniques. Recently, robust graph-based methods for NLP have also been gaining a lot of interest, e.g., in word clustering (Brew & im Walde, 2002) and prepositional phrase attachment (Toutanova, Manning, & Ng, 2004).

In this paper, we will take graph-based methods in NLP one step further. We will discuss how random walks on sentence-based graphs can help in text summarization. We will also briefly discuss how similar techniques can be applied to other NLP tasks such as named entity classification, prepositional phrase attachment, and text classification (e.g., spam recognition).

Text summarization is the process of automatically creating a compressed version of a given text that provides useful information for the user. The information content of a summary depends on user's needs. Topic-oriented summaries focus on a user's topic of interest, and extract the information in the text that is related to the specified topic. On the other hand, generic summaries try to cover as much of the information content as possible, preserving the general topical organization of the original text. In this paper, we focus on multi-document extractive generic text summarization, where the goal is to produce a summary of multiple documents about the same, but unspecified topic.

Extractive summarization produces summaries by choosing a subset of the sentences in the original document(s). This contrasts with abstractive summarization, where the information in the text is rephrased. Although summaries produced by humans are typically not extractive, most of the summarization research today is on extractive summarization. Purely extractive summaries often give better results compared to automatic abstractive summaries. This is due to the fact that the problems in abstractive summarization, such as semantic representation, inference and natural language generation, are relatively harder compared to a data-driven approach such as sentence extraction. In fact, truly abstractive summarization has not reached to a mature stage today. Existing abstractive summarizers often depend on an extractive preprocessing component. The output of the extractor is cut and pasted, or compressed to produce the abstract of the text (Witbrock & Mittal, 1999; Jing, 2002; Knight & Marcu, 2000). SUMMONS (Radev & McKeown, 1998) is an example of a multi-document summarizer which extracts and combines information from multiple sources and passes this information to a language generation component to produce the final summary.

Early research on extractive summarization is based on simple heuristic features of the sentences such as their position in the text, the overall frequency of the words they contain, or some key phrases indicating the importance of the sentences (Baxendale, 1958; Edmundson, 1969; Luhn, 1958). A commonly used measure to assess the importance of the words in a sentence is the *inverse document frequency*, or idf, which is defined by the formula (Sparck-Jones, 1972):

$$\mathrm{idf}_i = \log\Big(\frac{N}{n_i}\Big) \tag{1}$$

where $N$ is the total number of the documents in a collection, and $n_i$ is the number of documents in which word $i$ occurs. For example, the words that are likely to occur in almost every document (e.g. articles "a" and "the") have idf values close to zero while rare words (e.g. medical terms, proper nouns) typically have higher idf values.

More advanced techniques also consider the relation between sentences or the discourse structure by using synonyms of the words or anaphora resolution (Mani & Bloedorn, 1997; Barzilay & Elhadad, 1999). Researchers have also tried to integrate machine learning into summarization as more features have been proposed and more training data have become available (Kupiec, Pedersen, & Chen, 1995; Lin, 1999; Osborne, 2002; Daumé III & Marcu, 2004).

Our summarization approach in this paper is to assess the *centrality* of each sentence in a cluster and extract the most important ones to include in the summary. We investigate different ways of defining the lexical centrality principle in multi-document summarization, which measures centrality in terms of lexical properties of the sentences.

In Section 2, we present centroid-based summarization, a well-known method for judging sentence centrality. Then we introduce three new measures for centrality, Degree, LexRank with threshold, and continuous LexRank, inspired from the "prestige" concept in social networks. We propose a graph representation of a document cluster, where vertices represent the sentences and edges are defined in terms of the similarity relation between pairs of sentences. This representation enables us to make use of several centrality heuristics defined on graphs. We compare our new methods with centroid-based summarization using a feature-based generic summarization toolkit, MEAD, and show that our new features outperform Centroid in most of the cases. Test data for our experiments are taken from 2003 and 2004 summarization evaluations of Document Understanding Conferences (DUC) to compare our system with other state-of-the-art summarization systems and human performance as well.

## 2. Sentence Centrality and Centroid-based Summarization

Extractive summarization works by choosing a subset of the sentences in the original documents. This process can be viewed as identifying the most *central* sentences in a (multi-document) cluster that give the necessary and sufficient amount of information related to the main theme of the cluster. Centrality of a sentence is often defined in terms of the centrality of the words that it contains. A common way of assessing word centrality is to look at the centroid of the document cluster in a vector space. The centroid of a cluster is a pseudo-document which consists of words that have tf×idf scores above a predefined threshold, where tf is the frequency of a word in the cluster, and idf values are typically computed over a much larger and similar genre data set. In centroid-based summarization (Radev, Jing, & Budzikowska, 2000), the sentences that contain more words from the centroid of the cluster are considered as central (Algorithm 1). This is a measure of how close the sentence is to the centroid of the cluster. Centroid-based summarization has given promising results in the past, and it has resulted in the first web-based multi-document summarization system[1] (Radev, Blair-Goldensohn, & Zhang, 2001).

## 3. Centrality-based Sentence Salience

In this section, we propose several other criteria to assess sentence salience. All of our approaches are based on the concept of *prestige*[2] in social networks, which has also inspired many ideas in computer networks and information retrieval. A social network is a mapping of relationships between interacting entities (e.g. people, organizations, computers). Social networks are represented as graphs, where the nodes represent the entities and the links represent the relations between the nodes.

A cluster of documents can be viewed as a network of sentences that are related to each other. Some sentences are more similar to each other while some others may share only a little information with the rest of the sentences. We hypothesize that the sentences that are similar to many of the other sentences in a cluster are more central (or *salient*) to the topic. There are two points to clarify in this definition of centrality. First is how to

---

1. http://www.newsinessence.com
2. "Prestige" and "centrality" stand for the same concept with the difference that the former is often defined for directed graphs whereas the latter is defined for undirected graphs.

```
    input  : An array S of n sentences, cosine threshold t
    output : An array C of Centroid scores
 1  Hash WordHash;
 2  Array C;
 3  /* compute tf×idf scores for each word */
 4  for i ← 1 to n do
 5      foreach word w of S[i] do
 6          WordHash{w}{"tfidf"} = WordHash{w}{"tfidf"} + idf{w};
 7      end
 8  end

 9  /* construct the centroid of the cluster */
10  /* by taking the words that are above the threshold*/
11  foreach word w of WordHash do
12      if WordHash{w}{"tfidf"} > t then
13          WordHash{w}{"centroid"} = WordHash{w}{"tfidf"};
14      end
15      else
16          WordHash{w}{"centroid"} = 0;
17      end
18  end

19  /* compute the score for each sentence */
20  for i ← 1 to n do
21      C[i] = 0;
22      foreach word w of S[i] do
23          C[i] = C[i] + WordHash{w}{"centroid"};
24      end
25  end
26  return C;
```

**Algorithm 1**: Computing Centroid scores.

define similarity between two sentences. Second is how to compute the overall centrality of a sentence given its similarity to other sentences.

To define similarity, we use the bag-of-words model to represent each sentence as an $N$-dimensional vector, where $N$ is the number of all possible words in the target language. For each word that occurs in a sentence, the value of the corresponding dimension in the vector representation of the sentence is the number of occurrences of the word in the sentence times the idf of the word. The similarity between two sentences is then defined by the cosine between two corresponding vectors:

$$\text{idf-modified-cosine}(x,y) = \frac{\sum_{w \in x,y} \text{tf}_{w,x} \text{tf}_{w,y} (\text{idf}_w)^2}{\sqrt{\sum_{x_i \in x} (\text{tf}_{x_i,x} \text{idf}_{x_i})^2} \times \sqrt{\sum_{y_i \in y} (\text{tf}_{y_i,y} \text{idf}_{y_i})^2}} \tag{2}$$

where $\text{tf}_{w,s}$ is the number of occurrences of the word $w$ in the sentence $s$.

A cluster of documents may be represented by a cosine similarity matrix where each entry in the matrix is the similarity between the corresponding sentence pair. Figure 1 shows a subset of a cluster used in DUC 2004, and the corresponding cosine similarity matrix. Sentence ID d$X$s$Y$ indicates the $Y^{\text{th}}$ sentence in the $X^{\text{th}}$ document. This matrix can also be represented as a weighted graph where each edge shows the cosine similarity between a pair of sentence (Figure 2). In the following sections, we discuss several ways of computing sentence centrality using the cosine similarity matrix and the corresponding graph representation.

### 3.1 Degree Centrality

In a cluster of related documents, many of the sentences are expected to be somewhat similar to each other since they are all about the same topic. This can be seen in Figure 1 where the majority of the values in the similarity matrix are nonzero. Since we are interested in *significant* similarities, we can eliminate some low values in this matrix by defining a threshold so that the cluster can be viewed as an (undirected) graph, where each sentence of the cluster is a node, and significantly similar sentences are connected to each other. Figure 3 shows the graphs that correspond to the adjacency matrices derived by assuming the pair of sentences that have a similarity above 0.1, 0.2, and 0.3, respectively, in Figure 1 are similar to each other. Note that there should also be self links for all of the nodes in the graphs since every sentence is trivially similar to itself. Although we omit the self links for readability, the arguments in the following sections assume that they exist.

A simple way of assessing sentence centrality by looking at the graphs in Figure 3 is to count the number of similar sentences for each sentence. We define *degree centrality* of a sentence as the degree of the corresponding node in the similarity graph. As seen in Table 1, the choice of cosine threshold dramatically influences the interpretation of centrality. Too low thresholds may mistakenly take weak similarities into consideration while too high thresholds may lose many of the similarity relations in a cluster.

| ID | Degree (0.1) | Degree (0.2) | Degree (0.3) |
|---|---|---|---|
| d1s1 | 5 | 4 | 2 |
| d2s1 | 7 | 4 | 2 |
| d2s2 | 2 | 1 | 1 |
| d2s3 | 6 | 3 | 1 |
| d3s1 | 5 | 2 | 1 |
| d3s2 | 7 | 5 | 1 |
| d3s3 | 2 | 2 | 1 |
| d4s1 | 9 | 6 | 1 |
| d5s1 | 5 | 4 | 2 |
| d5s2 | 6 | 4 | 1 |
| d5s3 | 5 | 2 | 2 |

Table 1: Degree centrality scores for the graphs in Figure 3. Sentence d4s1 is the most central sentence for thresholds 0.1 and 0.2.

### 3.2 Eigenvector Centrality and LexRank

When computing degree centrality, we have treated each edge as a *vote* to determine the overall centrality value of each node. This is a totally democratic method where each vote counts the same. However, in many types of social networks, not all of the relationships are considered equally important. As an example, consider a social network of people that are connected to each other with the friendship relation. The prestige of a person does not only depend on how many friends he has, but also depends on *who* his friends are.

The same idea can be applied to extractive summarization as well. Degree centrality may have a negative effect in the quality of the summaries in some cases where several unwanted sentences vote for each other and raise their centrality. As an extreme example, consider a noisy cluster where all the documents are related to each other, but only one of them is about a somewhat different topic. Obviously, we would not want any of the sentences in the unrelated document to be included in a generic summary of the cluster. However, suppose that the unrelated document contains some sentences that are very prestigious considering only the votes in that document. These sentences will get artificially high centrality scores by the local votes from a specific set of sentences. This situation can be avoided by considering where the votes come from and taking the centrality of the *voting* nodes into account in weighting each vote. A straightforward way of formulating this idea is to consider every node having a centrality value and distributing this centrality to its neighbors. This formulation can be expressed by the equation

$$p(u) = \sum_{v \in adj[u]} \frac{p(v)}{deg(v)} \tag{3}$$

where $p(u)$ is the centrality of node $u$, $adj[u]$ is the set of nodes that are adjacent to $u$, and $deg(v)$ is the degree of the node $v$. Equivalently, we can write Equation 3 in the matrix notation as

$$\mathbf{p} = \mathbf{B}^\mathrm{T}\mathbf{p} \tag{4}$$

or

$$\mathbf{p}^\mathrm{T}\mathbf{B} = \mathbf{p}^\mathrm{T} \tag{5}$$

where the matrix $\mathbf{B}$ is obtained from the adjacency matrix of the similarity graph by dividing each element by the corresponding row sum:

$$\mathbf{B}(i,j) = \frac{\mathbf{A}(i,j)}{\sum_k \mathbf{A}(i,k)} \tag{6}$$

| SNo | ID | Text |
|---|---|---|
| 1 | d1s1 | Iraqi Vice President Taha Yassin Ramadan announced today, Sunday, that Iraq refuses to back down from its decision to stop cooperating with disarmament inspectors before its demands are met. |
| 2 | d2s1 | Iraqi Vice president Taha Yassin Ramadan announced today, Thursday, that Iraq rejects cooperating with the United Nations except on the issue of lifting the blockade imposed upon it since the year 1990. |
| 3 | d2s2 | Ramadan told reporters in Baghdad that "Iraq cannot deal positively with whoever represents the Security Council unless there was a clear stance on the issue of lifting the blockade off of it. |
| 4 | d2s3 | Baghdad had decided late last October to completely cease cooperating with the inspectors of the United Nations Special Commission (UNSCOM), in charge of disarming Iraq's weapons, and whose work became very limited since the fifth of August, and announced it will not resume its cooperation with the Commission even if it were subjected to a military operation. |
| 5 | d3s1 | The Russian Foreign Minister, Igor Ivanov, warned today, Wednesday, against using force against Iraq, which will destroy, according to him, seven years of difficult diplomatic work and will complicate the regional situation in the area. |
| 6 | d3s2 | Ivanov contended that carrying out air strikes against Iraq, who refuses to cooperate with the United Nations inspectors, "will end the tremendous work achieved by the international group during the past seven years and will complicate the situation in the region." |
| 7 | d3s3 | Nevertheless, Ivanov stressed that Baghdad must resume working with the Special Commission in charge of disarming the Iraqi weapons of mass destruction (UNSCOM). |
| 8 | d4s1 | The Special Representative of the United Nations Secretary-General in Baghdad, Prakash Shah, announced today, Wednesday, after meeting with the Iraqi Deputy Prime Minister Tariq Aziz, that Iraq refuses to back down from its decision to cut off cooperation with the disarmament inspectors. |
| 9 | d5s1 | British Prime Minister Tony Blair said today, Sunday, that the crisis between the international community and Iraq "did not end" and that Britain is still "ready, prepared, and able to strike Iraq." |
| 10 | d5s2 | In a gathering with the press held at the Prime Minister's office, Blair contended that the crisis with Iraq "will not end until Iraq has absolutely and unconditionally respected its commitments" towards the United Nations. |
| 11 | d5s3 | A spokesman for Tony Blair had indicated that the British Prime Minister gave permission to British Air Force Tornado planes stationed in Kuwait to join the aerial bombardment against Iraq. |

|    | 1 | 2 | 3 | 4 | 5 | 6 | 7 | 8 | 9 | 10 | 11 |
|----|---|---|---|---|---|---|---|---|---|----|----|
| 1  | 1.00 | 0.45 | 0.02 | 0.17 | 0.03 | 0.22 | 0.03 | 0.28 | 0.06 | 0.06 | 0.00 |
| 2  | 0.45 | 1.00 | 0.16 | 0.27 | 0.03 | 0.19 | 0.03 | 0.21 | 0.03 | 0.15 | 0.00 |
| 3  | 0.02 | 0.16 | 1.00 | 0.03 | 0.00 | 0.01 | 0.03 | 0.04 | 0.00 | 0.01 | 0.00 |
| 4  | 0.17 | 0.27 | 0.03 | 1.00 | 0.01 | 0.16 | 0.28 | 0.17 | 0.00 | 0.09 | 0.01 |
| 5  | 0.03 | 0.03 | 0.00 | 0.01 | 1.00 | 0.29 | 0.05 | 0.15 | 0.20 | 0.04 | 0.18 |
| 6  | 0.22 | 0.19 | 0.01 | 0.16 | 0.29 | 1.00 | 0.05 | 0.29 | 0.04 | 0.20 | 0.03 |
| 7  | 0.03 | 0.03 | 0.03 | 0.28 | 0.05 | 0.05 | 1.00 | 0.06 | 0.00 | 0.00 | 0.01 |
| 8  | 0.28 | 0.21 | 0.04 | 0.17 | 0.15 | 0.29 | 0.06 | 1.00 | 0.25 | 0.20 | 0.17 |
| 9  | 0.06 | 0.03 | 0.00 | 0.00 | 0.20 | 0.04 | 0.00 | 0.25 | 1.00 | 0.26 | 0.38 |
| 10 | 0.06 | 0.15 | 0.01 | 0.09 | 0.04 | 0.20 | 0.00 | 0.20 | 0.26 | 1.00 | 0.12 |
| 11 | 0.00 | 0.00 | 0.00 | 0.01 | 0.18 | 0.03 | 0.01 | 0.17 | 0.38 | 0.12 | 1.00 |

Figure 1: Intra-sentence cosine similarities in a subset of cluster d1003t from DUC 2004. Source: Agence France Presse (AFP) Arabic Newswire (1998). Manually translated to English.

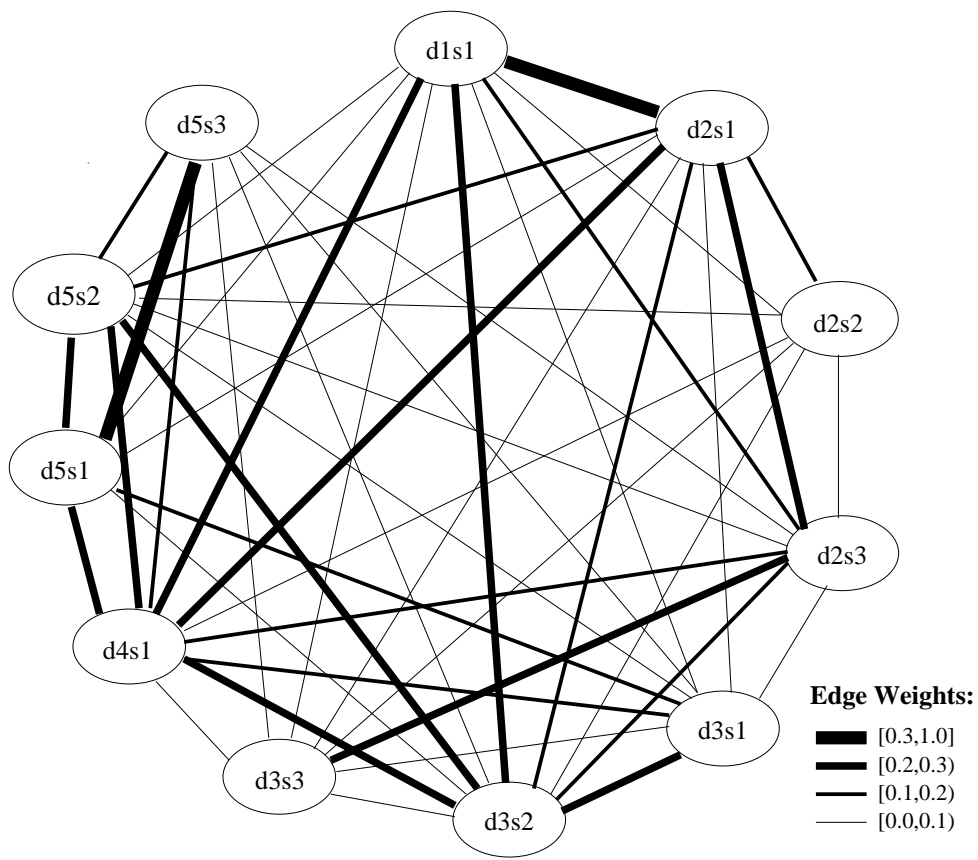

Figure 2: Weighted cosine similarity graph for the cluster in Figure 1.

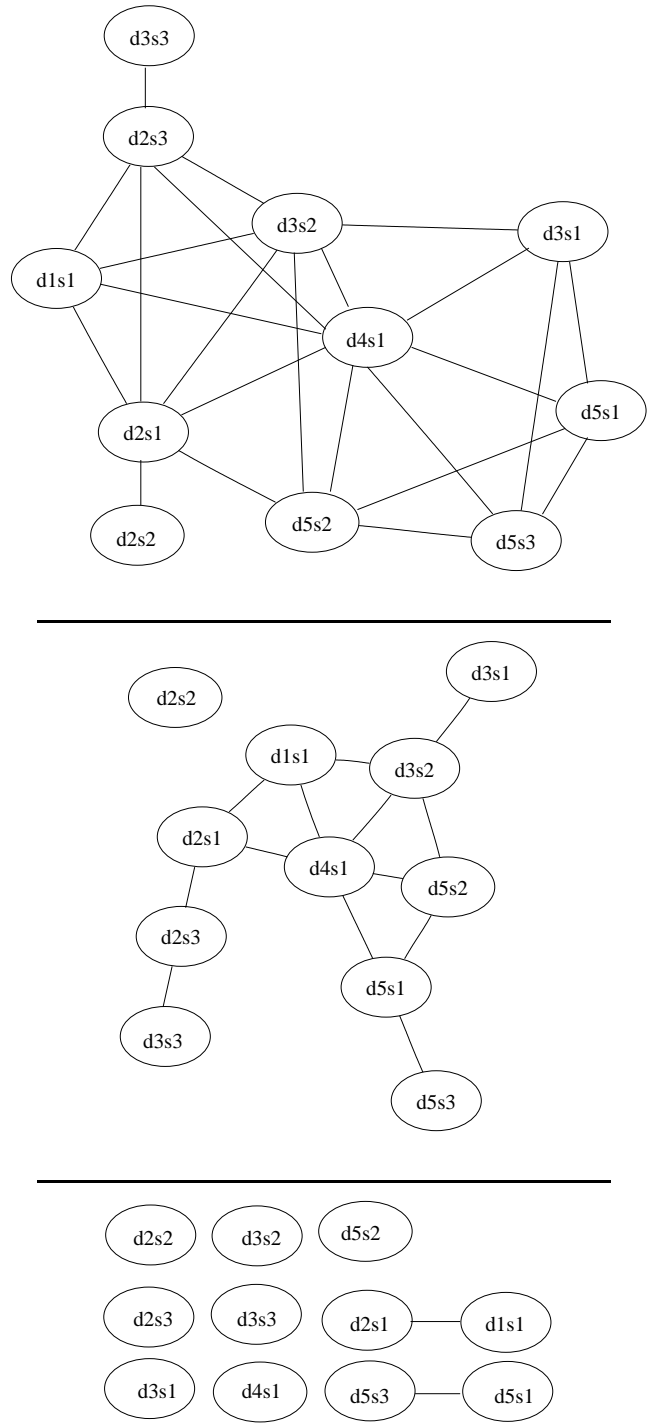

Figure 3: Similarity graphs that correspond to thresholds 0.1, 0.2, and 0.3, respectively, for the cluster in Figure 1.

Note that a row sum is equal to the degree of the corresponding node. Since every sentence is similar at least to itself, all row sums are nonzero. Equation 5 states that $\mathbf{p}^\mathrm{T}$ is the left eigenvector of the matrix $\mathbf{B}$ with the corresponding eigenvalue of 1. To guarantee that such an eigenvector exists and can be uniquely identified and computed, we need some mathematical foundations.

A *stochastic* matrix, $\mathbf{X}$, is the transition matrix of a Markov chain. An element $\mathbf{X}(i,j)$ of a stochastic matrix specifies the transition probability from state $i$ to state $j$ in the corresponding Markov chain. By the probability axioms, all rows of a stochastic matrix should add up to 1. $\mathbf{X^n}(i,j)$ gives the probability of reaching from state $i$ to state $j$ in $n$ transitions. A Markov chain with the stochastic matrix $\mathbf{X}$ converges to a stationary distribution if

$$\lim_{n \to \infty} \mathbf{X^n} = \mathbf{1}^\mathrm{T}\mathbf{r} \tag{7}$$

where $\mathbf{1} = (1, 1, ..., 1)$, and the vector $\mathbf{r}$ is called the stationary distribution of the Markov chain. An intuitive interpretation of the stationary distribution can be understood by the concept of a random walk. Each element of the vector $\mathbf{r}$ gives the asymptotic probability of ending up in the corresponding state in the long run regardless of the starting state. A Markov chain is *irreducible* if any state is reachable from any other state, i.e. for all $i, j$ there exists an $n$ such that $\mathbf{X^n}(i,j) \neq 0$. A Markov chain is *aperiodic* if for all $i$, $gcd\{n : X^n(i,i) > 0\} = 1$. By the Perron-Frobenius theorem (Seneta, 1981), an irreducible and aperiodic Markov chain is guaranteed to converge to a unique stationary distribution.

If a Markov chain has reducible or periodic components, a random walker may get stuck in these components and never visit the other parts of the graph.

Since the similarity matrix $\mathbf{B}$ in Equation 4 satisfies the properties of a stochastic matrix, we can treat it as a Markov chain. The centrality vector $\mathbf{p}$ corresponds to the stationary distribution of $\mathbf{B}$. However, we need to make sure that the similarity matrix is always irreducible and aperiodic. To solve this problem, Page et al. (1998) suggest reserving some low probability for jumping to any node in the graph. This way the random walker can "escape" from periodic or disconnected components, which makes the graph irreducible and aperiodic. If we assign a uniform probability for jumping to any node in the graph, we are left with the following modified version of Equation 3, which is known as PageRank,

$$p(u) = \frac{d}{N} + (1 - d) \sum_{v \in adj[u]} \frac{p(v)}{deg(v)} \qquad (8)$$

where $N$ is the total number of nodes in the graph, and $d$ is a "damping factor", which is typically chosen in the interval $[0.1, 0.2]$ (Brin & Page, 1998). Equation 8 can be written in the matrix form as

$$\mathbf{p} = [d\mathbf{U} + (1 - d)\mathbf{B}]^{\mathrm{T}} \mathbf{p} \qquad (9)$$

where $\mathbf{U}$ is a square matrix with all elements being equal to $1/N$. The transition kernel $[d\mathbf{U} + (1 - d)\mathbf{B}]$ of the resulting Markov chain is a mixture of two kernels $\mathbf{U}$ and $\mathbf{B}$. A random walker on this Markov chain chooses one of the adjacent states of the current state with probability $1 - d$, or jumps to any state in the graph, including the current state, with probability $d$. The PageRank formula was first proposed for computing web page prestige, and still serves as the underlying mechanism behind the Google search engine.

The convergence property of Markov chains also provides us with a simple iterative algorithm, called power method, to compute the stationary distribution (Algorithm 2). The algorithm starts with a uniform distribution. At each iteration, the eigenvector is updated by multiplying with the transpose of the stochastic matrix. Since the Markov chain is irreducible and aperiodic, the algorithm is guaranteed to terminate.

**input** : A stochastic, irreducible and aperiodic matrix $\mathbf{M}$
**input** : matrix size $N$, error tolerance $\epsilon$
**output**: eigenvector $\mathbf{p}$
1  $\mathbf{p}_0 = \frac{1}{N}\mathbf{1}$;
2  t=0;
3  **repeat**
4      t=t+1;
5      $\mathbf{p}_t = \mathbf{M}^{\mathrm{T}}\mathbf{p}_{t-1}$;
6      $\delta = ||\mathbf{p}_t - \mathbf{p}_{t-1}||$;
7  **until** $\delta < \epsilon$;
8  **return** $\mathbf{p}_t$;

**Algorithm 2**: Power Method for computing the stationary distribution of a Markov chain.

Unlike the original PageRank method, the similarity graph for sentences is undirected since cosine similarity is a symmetric relation. However, this does not make any difference in the computation of the stationary distribution. We call this new measure of sentence similarity *lexical PageRank*, or *LexRank*. Algorithm 3 summarizes how to compute LexRank scores for a given set of sentences. Note that Degree centrality scores are also computed (in the *Degree* array) as a side product of the algorithm. Table 2 shows the LexRank scores for the graphs in Figure 3 setting the damping factor to 0.85. For comparison, Centroid score for each sentence is also shown in the table. All the numbers are normalized so that the highest ranked sentence gets the score 1. It is obvious from the figures that threshold choice affects the LexRank rankings of some sentences.

```
1  MInput array S of n sentences, cosine threshold t   output: An array L of LexRank scores
2  Array CosineMatrix[n][n];
3  Array Degree[n];
4  Array L[n];
5  for i ← 1 to n do
6      for j ← 1 to n do
7          CosineMatrix[i][j] = idf-modified-cosine(S[i],S[j]);
8          if CosineMatrix[i][j] > t then
9              CosineMatrix[i][j] = 1;
10             Degree[i] + +;
11         end
12         else
13             CosineMatrix[i][j] = 0;
14         end
15     end
16 end
17 for i ← 1 to n do
18     for j ← 1 to n do
19         CosineMatrix[i][j] = CosineMatrix[i][j]/Degree[i];
20     end
21 end
22 L = PowerMethod(CosineMatrix,n,ε);
23 return L;
```

**Algorithm 3**: Computing LexRank scores.

| ID | LR (0.1) | LR (0.2) | LR (0.3) | Centroid |
|----|----------|----------|----------|----------|
| d1s1 | 0.6007 | 0.6944 | 1.0000 | 0.7209 |
| d2s1 | 0.8466 | 0.7317 | 1.0000 | 0.7249 |
| d2s2 | 0.3491 | 0.6773 | 1.0000 | 0.1356 |
| d2s3 | 0.7520 | 0.6550 | 1.0000 | 0.5694 |
| d3s1 | 0.5907 | 0.4344 | 1.0000 | 0.6331 |
| d3s2 | 0.7993 | 0.8718 | 1.0000 | 0.7972 |
| d3s3 | 0.3548 | 0.4993 | 1.0000 | 0.3328 |
| d4s1 | 1.0000 | 1.0000 | 1.0000 | 0.9414 |
| d5s1 | 0.5921 | 0.7399 | 1.0000 | 0.9580 |
| d5s2 | 0.6910 | 0.6967 | 1.0000 | 1.0000 |
| d5s3 | 0.5921 | 0.4501 | 1.0000 | 0.7902 |

Table 2: LexRank scores for the graphs in Figure 3. All the values are normalized so that the largest value of each column is 1. Sentence d4s1 is the most central page for thresholds 0.1 and 0.2.

## 3.3 Continuous LexRank

The similarity graphs we have constructed to compute Degree centrality and LexRank are unweighted. This is due to the binary discretization we perform on the cosine matrix using an appropriate threshold. As in all discretization operations, this means an information loss. One improvement over LexRank can be obtained by making use of the *strength* of the similarity links. If we use the cosine values directly to construct the similarity graph, we usually have a much denser but weighted graph (Figure 2). We can normalize the row sums of the corresponding transition matrix so that we have a stochastic matrix. The resultant equation is a modified version of LexRank for weighted graphs:

$$p(u) = \frac{d}{N} + (1 - d) \sum_{v \in adj[u]} \frac{\text{idf-modified-cosine}(u, v)}{\sum_{z \in adj[v]} \text{idf-modified-cosine}(z, v)} p(v) \tag{10}$$

This way, while computing LexRank for a sentence, we multiply the LexRank values of the linking sentences by the weights of the links. Weights are normalized by the row sums, and the damping factor $d$ is added for the convergence of the method.

## 3.4 Centrality vs. Centroid

Graph-based centrality has several advantages over Centroid. First of all, it accounts for information subsumption among sentences. If the information content of a sentence subsumes another sentence in a cluster, it is naturally preferred to include the one that contains more information in the summary. The degree of a node in the cosine similarity graph is an indication of how much common information the sentence has with other sentences. Sentence d4s1 in Figure 1 gets the highest score since it almost subsumes the information in the first two sentences of the cluster and has some common information with others. Another advantage of our proposed approach is that it prevents unnaturally high idf scores from boosting up the score of a sentence that is unrelated to the topic. Although the frequency of the words are taken into account while computing the Centroid score, a sentence that contains many rare words with high idf values may get a high Centroid score even if the words do not occur elsewhere in the cluster.

## 4. Experimental Setup

In this section, we describe the data set, the evaluation metric and the summarization system we used in our experiments.

### 4.1 Data Set and Evaluation Method

We used DUC 2003 and 2004 data sets in our experiments. Task 2 of both DUC 2003 and 2004 involve generic summarization of news documents clusters. There are a total of 30 clusters in DUC 2003 and 50 clusters in DUC 2004. In addition to these two tasks, we used two more data sets from Task 4 of DUC 2004, which involves cross-lingual generic summarization. First set (Task 4a) is composed of Arabic-to-English machine translations of 24 news clusters. Second set (Task 4b) is the human translations of the same clusters. All data sets are in English.

For evaluation, we used the new automatic summary evaluation metric, ROUGE[3], which was used for the first time in DUC 2004. ROUGE is a recall-based metric for fixed-length summaries which is based on n-gram co-occurrence. It reports separate scores for 1, 2, 3, and 4-gram matching between the model summaries and the summary to be evaluated. Among these different scores, unigram-based ROUGE score (ROUGE-1) has been shown to agree with human judgements most (Lin & Hovy, 2003).

---

3. `http://www.isi.edu/~cyl/ROUGE`

There are 10 different human judges for DUC 2003 Task 2; 8 for DUC 2004 Task 2; and 4 for DUC 2004 Task 4. However, a subset of exactly 4 different human judges produced model summaries for any given cluster. ROUGE requires a limit on the length of the summaries to be able to make a fair evaluation. To stick with the DUC 2004 specifications and to be able to compare our system with human summaries and as well as with other DUC participants, we produced 665-byte summaries for each cluster and computed ROUGE scores against human summaries.

## 4.2 MEAD Summarization Toolkit

We implemented our methods inside the MEAD[4] summarization system (Radev et al., 2001). MEAD is a publicly available toolkit for extractive multi-document summarization. Although it comes as a centroid-based summarization system by default, its feature set can be extended to implement any other method.

The MEAD summarizer consists of three components. During the first step, *the feature extraction*, each sentence in the input document (or cluster of documents) is converted into a feature vector using the user-defined features. Second, the feature vector is converted to a scalar value using the *combiner*. Combiner outputs a linear combination of the features by using the predefined feature weights. At the last stage known as the *reranker*, the scores for sentences included in related pairs are adjusted upwards or downwards based on the type of relation between the sentences in the pair. Reranker penalizes the sentences that are similar to the sentences already included in the summary so that a better information coverage is achieved.

Three default features that come with the MEAD distribution are Centroid, Position and Length. Position is the normalized value of the position of a sentence in the document such that the first sentence of a document gets the maximum Position value of 1, and the last sentence gets the value 0. Length is not a real feature score, but a cutoff value that ignores sentences shorter than the given threshold. Several rerankers are implemented in MEAD, including one based on Maximal Marginal Relevance (MMR) (Carbonell & Goldstein, 1998) and the default reranker of the system based on Cross-Sentence Informational Subsumption (CSIS) (Radev, 2000). All of our experiments shown in Section 5 use the CSIS reranker.

A MEAD policy is a combination of three components: (a) the command lines for all features, (b) the formula for converting the feature vector to a scalar, and (c) the command line for the reranker. A sample policy might be the one shown in Figure 4. This example indicates the three default MEAD features (Centroid, Position, LengthCutoff), and our new LexRank feature used in our experiments. Our LexRank implementation requires the cosine similarity threshold, 0.2 in the example, as an argument. Each number next to a feature name shows the relative weight of that feature (except for LengthCutoff where the number 9 indicates the threshold for selecting a sentence based on the number of the words in the sentence). The reranker in the example is a word-based MMR reranker with

---

4. http://www.summarization.com

```
feature LexRank LexRank.pl 0.2
Centroid 1 Position 1 LengthCutoff 9 LexRank 1
mmr-reranker-word.pl 0.5 MEAD-cosine enidf
```

Figure 4: A sample MEAD policy.

a cosine similarity threshold, 0.5. Finally "enidf" specifies the idf database file, which is a precomputed list of idf's for English words.

## 5. Results and Discussion

The following sections show the results of the experiments we have performed on the official DUC data sets with different implementations of similarity graph based centrality. We have implemented Degree centrality, LexRank with threshold and continuous LexRank as separate features in MEAD. All the feature values are normalized so that the sentence that has the highest value gets the score 1, and the sentence with the lowest value gets the score 0. In all of the runs, we have used Length and Position features of MEAD as supporting heuristics in addition to our centrality features. Length cutoff value is set to 9, i.e. all the sentences that have less than 9 words are discarded. The weight of the Position feature is fixed to 1 in all runs. Other than these two heuristic features, we used each centrality feature alone without combining with other centrality methods to make a better comparison with each other. For each centrality feature we are experimenting with, we have run 8 different MEAD features by setting the weight of the corresponding feature to $0.5, 1.0, 1.5, 2.0, 2.5, 3.0, 5.0, 10.0$, respectively.

### 5.1 Effect of Threshold on Degree and LexRank Centrality

We have demonstrated that very high thresholds may lose almost all of the information in a similarity matrix (Figure 3). To support our claim, we have run Degree and LexRank centrality with different thresholds for our data sets. Figure 5 shows the effect of threshold for Degree and LexRank centrality on DUC 2004 Task 2 data. We have experimented with four different thresholds: 0.1, 0.2, 0.3, and 0.4. Eight different data points shown for each threshold correspond to the runs of the same feature with eight different weights as we have discussed above. The mean value of the eight different experiments is shown as a horizontal line. It is apparent in the figures that the lowest threshold, 0.1, produces the best summaries. This means that the information loss in higher thresholds is high enough to result in worse ROUGE scores. The loss in ROUGE scores as we move from threshold 0.1 to 0.2 is more significant in Degree centrality.

This effect of threshold is an indication that our new centrality methods actually *work* for extractive summarization. The higher the threshold, the less informative, or even misleading, similarity graphs we must have. On the extreme point where we have a very high threshold, we would have no edges in the graph so that Degree or LexRank centrality would be of no use.

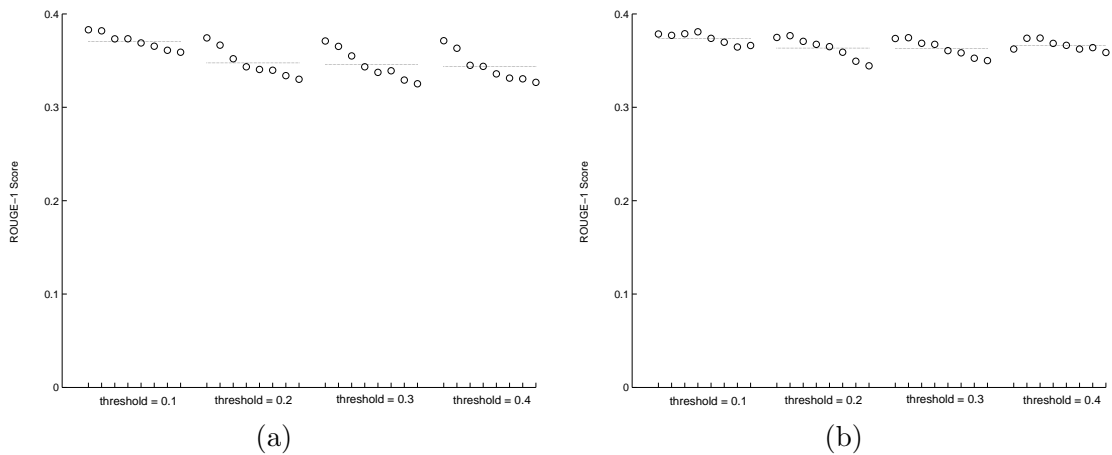

Figure 5: ROUGE-1 scores for (a) Degree centrality and (b) LexRank centrality with different thresholds on DUC 2004 Task 2 data.

## 5.2 Comparison of Centrality Methods

Table 3 shows the ROUGE scores for our experiments on DUC 2003 Task 2, DUC 2004 Task 2, DUC 2004 Task 4a, and DUC 2004 Task 4b, respectively. We show the minimum, the maximum, and the average ROUGE-1 scores for eight experiments we have run for each centrality method corresponding to eight different feature weights we have mentioned in Section 5. We include Degree and LexRank experiments only with threshold 0.1, the best one we have observed. We also include two baselines for each data set. The first baseline we have used is extracting random sentences from the cluster. We have performed five random runs for each data set. The results in the tables are for the median runs. The second baseline, shown as 'lead-based' in the tables, is using only the Position feature without any centrality method. This is tantamount to producing lead-based summaries, which is a widely used and very challenging baseline in the text summarization community (Brandow, Mitze, & Rau, 1995).

The top scores we have got in all data sets come from our new methods. All of our three new methods (Degree, LexRank with threshold, and continuous LexRank) perform significantly better than the baselines in all data sets. They also perform better than centroid-based summaries except for the DUC 2003 data set where the difference between Centroid and the others is not obvious. 0.1 seems to be an appropriate threshold such that the results seem as successful as using continuous LexRank. It is also hard to say that Degree and LexRank are significantly different from each other. This is an indication that Degree may already be a good enough measure to assess the centrality of a node in the similarity graph. Considering the relatively low complexity of degree centrality, it still serves as a plausible alternative when one needs a simple implementation. Computation of Degree can be done on the fly as a side product of LexRank just before the power method is applied on the similarity graph.

To have an idea of the relative success of our methods among other summarization systems, we have compared our ROUGE scores with other participants' scores in the same

DUC data sets. Table 4 and Table 5 show the official ROUGE-1 scores for top five participants and human summarizers on DUC 2003 and 2004 data, respectively. Most of the LexRank scores we got are better than the second best system in DUC 2003 and worse than the best system. Best few scores for each method are always statistically indistinguishable from the best system in the official evaluations considering the 95% confidence interval. On all three DUC 2004 data sets, we achieved a better score than the best participant in at least one of the policies we tried. On the DUC 2003 data, we achieved several scores that are between the best and the second best system.

|  | 2003 Task2 | | |
|---|---|---|---|
|  | min | max | average |
| Centroid | 0.3523 | 0.3713 | 0.3624 |
| Degree (t=0.1) | 0.3566 | 0.3640 | 0.3595 |
| LexRank (t=0.1) | 0.3610 | 0.3726 | 0.3666 |
| Cont. LexRank | 0.3594 | 0.3700 | 0.3646 |

baselines:    random:    0.3261
              lead-based:    0.3575

(a)

|  | 2004 Task2 | | |
|---|---|---|---|
|  | min | max | average |
| Centroid | 0.3580 | 0.3767 | 0.3670 |
| Degree (t=0.1) | 0.3590 | 0.3830 | 0.3707 |
| LexRank (t=0.1) | 0.3646 | 0.3808 | 0.3736 |
| Cont. LexRank | 0.3617 | 0.3826 | 0.3758 |

baselines:    random:    0.3238
              lead-based:    0.3686

(b)

|  | 2004 Task4a | | |
|---|---|---|---|
|  | min | max | average |
| Centroid | 0.3768 | 0.3901 | 0.3826 |
| Degree (t=0.1) | 0.3863 | 0.4027 | 0.3928 |
| LexRank (t=0.1) | 0.3931 | 0.4038 | 0.3974 |
| Cont. LexRank | 0.3924 | 0.4002 | 0.3963 |

baselines:    random:    0.3593
              lead-based:    0.3788

(c)

|  | 2004 Task4b | | |
|---|---|---|---|
|  | min | max | average |
| Centroid | 0.3760 | 0.3962 | 0.4034 |
| Degree (t=0.1) | 0.3801 | 0.4147 | 0.4026 |
| LexRank (t=0.1) | 0.3837 | 0.4167 | 0.4052 |
| Cont. LexRank | 0.3772 | 0.4082 | 0.3966 |

baselines:    random:    0.3734
              lead-based:    0.3587

(d)

Table 3: ROUGE-1 scores for different MEAD policies on DUC 2003 and 2004 data.

## 5.3 Experiments on Noisy Data

The graph-based methods we have proposed consider a document cluster as a whole. The centrality of a sentence is measured by looking at the overall interaction of the sentence within the cluster rather than the local value of the sentence in its document. This is especially critical in generic summarization where the information unrelated to the main theme of the cluster should be excluded from the summary. DUC data sets are perfectly clustered into related documents by human assessors. To observe the behavior of our methods on noisy data, we have added 2 random documents in each cluster taken from a different cluster. Since originally each cluster contains 10 documents, this means a 2/12 (17%) noise on the data sets.

The results on the noisy data are given in Table 6. The picture looks similar to Table 3 except for lead-based and random baselines are more significantly affected by the noise. The performance loss is quite small on our graph-based centrality methods. A surprising point is that centroid-based summarization also gives good results although still worse than the others most of the time. This suggests that 17% noise on the data is not enough to make significant changes on the centroid of a cluster.

| TASK 2 | | |
|---|---|---|
| **Peer Code** | **ROUGE-1 Score** | **95% Confidence Interval** |
| C | 0.4443 | [0.3924,0.4963] |
| B | 0.4425 | [0.4138,0.4711] |
| D | 0.4344 | [0.3821,0.4868] |
| E | 0.4218 | [0.3871,0.4565] |
| A | 0.4168 | [0.3864,0.4472] |
| I | 0.4055 | [0.3740,0.4371] |
| G | 0.3978 | [0.3765,0.4192] |
| F | 0.3904 | [0.3596,0.4211] |
| J | 0.3895 | [0.3591,0.4199] |
| H | 0.3869 | [0.3659,0.4078] |
| 12 | 0.3798 | [0.3598,0.3998] |
| 13 | 0.3676 | [0.3507,0.3844] |
| 16 | 0.3660 | [0.3474,0.3846] |
| 6 | 0.3607 | [0.3415,0.3799] |
| 26 | 0.3582 | [0.3337,0.3828] |

Table 4: Summary of official ROUGE scores for DUC 2003 Task 2. Peer codes: manual summaries [A-J] and top five system submissions.

| TASK 2 | | |
|---|---|---|
| **Peer Code** | **ROUGE-1 Score** | **95% Confidence Interval** |
| H | 0.4183 | [0.4019,0.4346] |
| F | 0.4125 | [0.3916,0.4333] |
| E | 0.4104 | [0.3882,0.4326] |
| D | 0.4059 | [0.3870,0.4249] |
| B | 0.4043 | [0.3795,0.4291] |
| A | 0.3933 | [0.3722,0.4143] |
| C | 0.3904 | [0.3715,0.4093] |
| G | 0.3890 | [0.3679,0.4101] |
| 65 | 0.3822 | [0.3694,0.3951] |
| 104 | 0.3744 | [0.3635,0.3853] |
| 35 | 0.3743 | [0.3612,0.3874] |
| 19 | 0.3739 | [0.3608,0.3869] |
| 124 | 0.3706 | [0.3578,0.3835] |

| TASK 4 | | |
|---|---|---|
| **Peer Code** | **ROUGE-1 Score** | **95% Confidence Interval** |
| Y | 0.4445 | [0.4230,0.4660] |
| Z | 0.4326 | [0.4088,0.4565] |
| X | 0.4293 | [0.4068,0.4517] |
| W | 0.4119 | [0.3870,0.4368] |
| Task 4a | | |
| 144 | 0.3883 | [0.3626,0.4139] |
| 22 | 0.3865 | [0.3635,0.4096] |
| 107 | 0.3862 | [0.3555,0.4168] |
| 68 | 0.3816 | [0.3642,0.3989] |
| 40 | 0.3796 | [0.3581,0.4011] |
| Task 4b | | |
| 23 | 0.4158 | [0.3933,0.4382] |
| 84 | 0.4101 | [0.3854,0.4348] |
| 145 | 0.4060 | [0.3678,0.4442] |
| 108 | 0.4006 | [0.3700,0.4312] |
| 69 | 0.3984 | [0.3744,0.4225] |

Table 5: Summary of official ROUGE scores for DUC 2004 Tasks 2 and 4. Peer codes: manual summaries [A-Z] and top five system submissions. Systems numbered 144 and 145 are University of Michigan's submission. 144 uses LexRank in combination with Centroid whereas 145 uses Centroid alone.

| | 2003 Task2 | | |
|---|---|---|---|
| | min | max | average |
| Centroid | 0.3502 | 0.3689 | 0.3617 |
| Degree (t=0.1) | 0.3501 | 0.3650 | 0.3573 |
| LexRank (t=0.1) | 0.3493 | 0.3677 | 0.3603 |
| Cont. LexRank | 0.3564 | 0.3653 | 0.3621 |

baselines:     random:     0.2952
           lead-based:   0.3246

(a)

| | 2004 Task2 | | |
|---|---|---|---|
| | min | max | average |
| Centroid | 0.3563 | 0.3732 | 0.3630 |
| Degree (t=0.1) | 0.3495 | 0.3762 | 0.3622 |
| LexRank (t=0.1) | 0.3512 | 0.3760 | 0.3663 |
| Cont. LexRank | 0.3465 | 0.3808 | 0.3686 |

baselines:     random:     0.3078
           lead-based:   0.3418

(b)

| | 2004 Task4a | | |
|---|---|---|---|
| | min | max | average |
| Centroid | 0.3706 | 0.3898 | 0.3761 |
| Degree (t=0.1) | 0.3874 | 0.3943 | 0.3906 |
| LexRank (t=0.1) | 0.3883 | 0.3992 | 0.3928 |
| Cont. LexRank | 0.3889 | 0.3931 | 0.3908 |

baselines:     random:     0.3315
           lead-based:   0.3615

(c)

| | 2004 Task4b | | |
|---|---|---|---|
| | min | max | average |
| Centroid | 0.3754 | 0.3942 | 0.3906 |
| Degree (t=0.1) | 0.3801 | 0.4090 | 0.3963 |
| LexRank (t=0.1) | 0.3710 | 0.4022 | 0.3911 |
| Cont. LexRank | 0.3700 | 0.4012 | 0.3905 |

baselines:     random:     0.3391
           lead-based:   0.3430

(d)

Table 6: ROUGE-1 scores for different MEAD policies on 17% noisy DUC 2003 and 2004 data.

## 6. Related Work

There have been attempts for using graph-based ranking methods in natural language applications before. Salton et al. (1997) made one of the first attempts of using degree centrality in single document text summarization. In the summarization approach of Salton et al., degree scores are used to extract the important paragraphs of a text.

Moens, Uyttendaele, and Dumortier (1999) use cosine similarity between the sentences to cluster a text into different topical regions. A predefined cosine threshold is used to cluster paragraphs around seed paragraphs (called *medoids*). Seed paragraphs are determined by maximizing the total similarity between the seed and the other paragraphs in a cluster. The seed paragraphs are then considered as the representative descriptions of the corresponding subtopics, and included in the summary.

Zha (2002) defines a bipartite graph from the set of terms to the set of sentences. There is an edge from a term $t$ to a sentence $s$ if $t$ occurs in $s$. Zha argues that the terms that appear in many sentences with high salience scores should have high salience scores, and the sentences that contain many terms with high salience scores should also have high salience scores. This *mutual reinforcement principal* reduces to a solution for the singular vectors of the transition matrix of the bipartite graph.

The work presented in this paper started with the implementation of LexRank with threshold on unweighted graphs. This implementation was first used in the DUC 2004 evaluations which was run in February 2004 and presented in May 2004 (Erkan & Radev, 2004b). After the DUC evaluations, a more detailed analysis and more careful implementation of the method was presented together with a comparison against degree centrality and centroid-based summarization (Erkan & Radev, 2004a). Continuous LexRank on weighted graphs first appeared in the initial version of this paper submitted in July 2004. An eigenvector centrality algorithm on weighted graphs was independently proposed by Mihalcea and Tarau (2004) for single-document summarization. Mihalcea, Tarau, and Figa (2004) later applied PageRank to another problem of natural language processing, word sense disambiguation.

Unlike our system, the studies mentioned above do not make use of any heuristic features of the sentences other than the centrality score. They do not also deal with the multi-document case. One of the main problems with multi-document summarization is the potential duplicate information coming from different documents, which is less likely to occur in single-document summaries. We try to avoid the repeated information in the summaries by using the reranker of the MEAD system. This problem is also addressed in Salton et al.'s work. Instead of using a reranker, they first segment the text into regions of different subtopics and then take at least one representative paragraph with the highest degree value from each region.

To determine the similarity between two sentences, we have used the cosine similarity metric that is based on word overlap and idf weighting. However, there are more advanced techniques of assessing similarity which are often used in the topical clustering of documents or sentences (Hatzivassiloglou et al., 2001; McKeown et al., 2001). The similarity computation might be improved by incorporating more features (e.g. synonym overlap, verb/argument structure overlap, stem overlap) or mechanisms (e.g. coreference resolution, paraphrasing) into the system. These improvements are orthogonal to our model in this paper and can be easily integrated into the similarity relation.

## 7. Conclusion

We have presented a new approach to define sentence salience based on graph-based centrality scoring of sentences. Constructing the similarity graph of sentences provides us with a better view of *important* sentences compared to the centroid approach, which is prone to over-generalization of the information in a document cluster. We have introduced three different methods for computing centrality in similarity graphs. The results of applying these methods on extractive summarization are quite promising. Even the simplest approach we have taken, degree centrality, is a good enough heuristic to perform better than lead-based and centroid-based summaries. In LexRank, we have tried to make use of more of the information in the graph, and got even better results in most of the cases. Lastly, we have shown that our methods are quite insensitive to noisy data that often occurs as a result of imperfect topical document clustering algorithms.

The graph-based representation of the relations between natural language constructs provides us with many new ways of information processing with applications to several problems such as document clustering, word sense disambiguation, prepositional phrase attachment. The similarity relation we used to construct the graphs can be replaced by any mutual information relation between natural language entities. We are currently working on using random walks on bipartite graphs (binary features on the left, objects to be classified on the right) for semi-supervised classification. For example, objects can be email messages and a binary feature may be "does the subject line of this message contain the word *money*". All objects are linked to the features that apply to them. A path through the graph can then go from an unlabeled object to a set of labeled ones going through a sequence of other objects and features. In traditional supervised or semi-supervised learning, one could not make effective use of the features solely associated with unlabeled examples. In this framework, these features serve as intermediate nodes on a path from unlabeled to labeled nodes. An eigenvector centrality method can then associate a probability with each object (labeled or unlabeled). That probability can then in turn be interpreted as belief in the classification of the object (e.g., there is an 87% per cent chance that this particular email message is spam). In an active learning setting, one can also choose what label to request next from an Oracle given the eigenvector centrality values of all objects.

## Acknowledgments

We would like to thank Mark Newman for providing some useful references for this paper. Thanks also go to Lillian Lee for her very helpful comments on an earlier version of this paper. Finally, we would like to thank the members of the CLAIR (Computational Linguistics And Information Retrieval) group at the University of Michigan, in particular Siwei Shen, for their assistance with this project.

This work was partially supported by the National Science Foundation under grant 0329043 "Probabilistic and link-based Methods for Exploiting Very Large Textual Repositories" administered through the IDM program. All opinions, findings, conclusions, and recommendations in this paper are made by the authors and do not necessarily reflect the views of the National Science Foundation.